\documentclass{article}

\usepackage[final,nonatbib]{neurips_2021}
\usepackage{titlesec}
\titlespacing*{\section}{0pt}{1.1\baselineskip}{\baselineskip}
\usepackage{amsmath}
\usepackage[utf8]{inputenc} 
\usepackage[T1]{fontenc}    
\usepackage{hyperref}       
\usepackage{url}            
\usepackage{booktabs}       
\usepackage{tabto} 
\usepackage{amsfonts}       
\usepackage{nicefrac}       
\usepackage{microtype}      
\usepackage{xcolor} 

\usepackage{algorithm} 
\usepackage{algpseudocode} 
\usepackage{enumitem}
\title{Simultaneous Improvement of ML Model Fairness and Performance by Identifying Bias in Data }

\author{%
  Aakash Agarwal, Bhushan Chaudhari, Dr. Tanmoy Bhowmik\\
  AI Garage Mastercard\\
 \texttt{(aakash.agarwal,bhushan.chaudhari,tanmoy.bhowmik)@mastercard.com} \\
}

\begin{document}

\maketitle

\begin{abstract}
  Machine learning models built on datasets containing discriminative instances attributed to various underlying factors result in biased and unfair outcomes. It’s a well founded and intuitive fact that existing bias mitigation strategies often sacrifice accuracy in order to ensure fairness. But when AI engine’s prediction is used for decision making which reflects on revenue or operational efficiency such as credit risk modeling, it would be desirable by the business if accuracy can be somehow reasonably preserved. This conflicting requirement of maintaining accuracy and fairness in AI motivates our research. In this paper, we propose a fresh approach for simultaneous improvement of fairness and accuracy of ML models within a realistic paradigm. The essence of our work is a data preprocessing technique that can detect instances ascribing a specific kind of bias that should be removed from the dataset before training and we further show that such instance removal will have no adverse impact on model accuracy. In particular, we claim that in the problem settings where instances exist with similar feature but different labels caused by variation in protected attributes, an inherent bias gets induced in the dataset, which can be identified and mitigated through our novel scheme. Our experimental evaluation on two open-source datasets demonstrates how the proposed method can mitigate bias along with  improving rather than degrading accuracy, while offering certain set of control for end user.
\end{abstract}

\section{Introduction}
AI powered predictive modeling techniques have been widely adopted by business verticals in different domains such as finance, healthcare, sports, banking, etc, often for making sensitive decisions ranging from personalized marketing, loan application approval (Mukerjee  et al. 2002) [3] to dating and hiring process(Bogen 2018, Cohen 2019)[4,5]. Unsurprisingly, with the continuous evolution and ever-increasing complexity, there have been several recent high-profile examples of machine learning (ML) going wrong in terms of bias, fairness and interpretability.

The presence of unintended demographic disparities or differential/disproportionate impact on individuals by machine learning models is demonstrated by (Calders 2013) [29]. Unfairness can be imparted in models because of bias present in training data. Various types of bias such as annotation bias, historical bias, prejudice bias, etc may lead to unfair models and selective bias towards a particular group. In (Mehrabi et al. 2019)[7], the authors have provided a comprehensive coverage of such biases supportaed by real life examples. 

The definitions used to understand the bias in models can be broadly categorized into three types: independence, separation and sufficiency (Sharma et al.2020)[10]. Specifically, a classifier satisfies independence if the protected attribute (such as race or gender) for which the model may be biased is independent of the classifier decision. Separation is satisfied if the classifier decision is independent of the protected attribute conditioned on the true label. Sufficiency is satisfied if the true label is independent of the protected attribute conditioned on the classifier prediction. Details on these fairness criteria, both mathematically and with respect to different worldviews, may be found in (Barocas 2019, Yeom 2018) [27,28] along with definitions of fairness metrics, such as statistical parity difference for independence and average odds difference for separation, from (Garg et al. 2020, Bellamy et al. 2019, Sharma et al. 2020) [8,9,10]. We are taking statistical parity difference and average odds difference metrics into consideration for this paper while being aware of the fact that there are various fairness metrics which are relevant to gauge biasness of models. Determining the right measure to be used must consider the proper legal, ethical, and social context.

Using these fairness metrics, several bias mitigation algorithms are developed to satisfy the various criteria of fairness for machine learning models to reduce bias. Methods to mitigate bias generally fall into three categories. Pre-processing techniques transform the data so that the underlying discrimination is removed (Alessandro  2017)[21]. If the algorithm is allowed to modify the training data, then pre-processing can be used (Bellamy et al. 2018)[22]. Our proposed methodology falls under this category as explained in the subsequent sections. In-processing techniques try to modify and change state-of-the-art learning algorithms in order to remove discrimination during the model training process (Alessandro  2017) [21]. If the algorithm can only treat the learned model as a black box without any ability to modify the training data or learning algorithm, then only post-processing can be used in which the labels assigned by the black-box model initially get reassigned based on a function during the post-processing phase (Bellamy et al. 2018)[22].

The literature extensively discusses the inherent trade-off between accuracy and fairness – as we pursue a higher degree of fairness, we may compromise accuracy (see for example (Kleinberg  et al. 2017) [12]). Many papers have empirically supported the existence of this trade-off (Be-chavod 2017, Friedler et al. 2019) [13, 14]. Generally, the aspiration of a fairness-aware algorithm is to develop a model that is fair without significantly compromising the accuracy or other alternative notions of utility.

In this paper, we are proposing a method to find bias inducing samples in the dataset and then dropping these samples such that the pre-processed dataset represents a more equitable world. In an equitable world, model outcome is independent of the protected attributes (such as gender, race, etc). Our novel approach can be described as follows: given a dataset that contains a protected attribute (such as gender, race, etc), samples with similar attributes but different protected attributes and different outcomes are flagged. For example – Credit Risk dataset contains 2 samples: 1 male and 1 female such that male and female sample have same attributes but model predicted low risk for male and high risk for female. We establish that these instances induce bias as the attributes are same but the outcome is different due to dependence on protected attributes (such as male, female) and thereby result in unfair treatment by the model. Further, protected attributes are not used for modelling, so such samples can confuse the model as they have nearly have the same attributes but different label and thus can be viewed as pseudo label noise. 

Hence our objective boils down to detect and remove such instances before training to make sure the resultant model is fairer. In the process we also show that such close instance removal does not compromise on the  model performance, rather on the contrary, it improves the accuracy. This simultaneous improvement of model fairness and accuracy which are in contrast of each other, although seems to be astonishing, but we could provide rationale for this achievement using prior work of Frénay et. al. [16]. This prior art explains the affect of such noisy instances on model performance and claim that label noise hampers the performance of the classifier which is also backed by (Long 2008)[15]. 

To summarize, in this paper, we make the following contributions : 
    \begin{enumerate}[noitemsep,nolistsep]
        \item We propose a systematic way of identification of bias inducing instances as per our definition in the previous section, and their subsequent removal from training data.
        \item We show that how the bias inducing instances removal ensures model fairness using the standard fairness metric.
        \item Further we show improvement in model accuracy trained on the bias eliminated data along with justification.
        \item We offer control in terms of adjustable hyperparameters to adjust fairness and accuracy as per the dataset and business requirements.
    \end{enumerate}
    
\section{Proposed Methodology}
We are proposing a method to filter out potential bias creating samples from the dataset based on certain similarity criteria. We present the complete methodology for the proposed solution in algorithm 1.
\hbadness=99999
\begin{algorithm}
\caption{Function to get Unbiased Data}
	
{\bf{\textit{GetUnbiasedData}}}(data, protected attribute , privileged group, label, favourable label)\newline
{\textbf{Input :}} Training data where protected attribute is one of the feature and label is target variable which will be used for prediction. \newline
{\textbf{Output  :}} Unbiased data/ Unbiased Model\newline
{\textbf{Steps  :}}\newline

•	Remove  correlated features with protected attribute.\newline
•	Normalize the continuous features and one hot encode the categorical features.\newline
•	Prepare two different groups based on protected attribute and output label\newline
\indent 1)Samples with privileged protected attribute and favourable output label\newline
\indent 2)Samples with un-privileged protected attribute and non-favourable output label\newline
•	Calculate cosine similarity between group 1 samples to group 2 samples.\newline
•	Flag similar samples from both groups based on cosine similarity threshold.\newline
•	Rank these similar samples (flagged from previous step) as per count of samples it is similar to with opposite group such that samples with higher count are at the top. \newline
•	Remove the top k\% similar samples from both groups.\newline
•	Apply reweighing technique on remaining instances to ensure same base rates with respect to protected attribute and output.\newline
•	Drop protected attribute and Fit any model of your choice.\newline

\end{algorithm}

All the respective evaluation and fairness metrics are calculated on the model outcomes of the test dataset. As discussed, hyperparameters – top k\% instances to remove and minimum similarity score are tuned to generate the results.

\section{Experiments}
This section describes the experiment design and performance of our proposed methodology when tested on two open-source datasets. Our focus here is on the improved fairness and accuracy obtained when the filtered dataset is used to train standard prediction algorithms. To perform the experiments, we have selected two datasets commonly used in the fairness research: UCI Adult dataset [1] and German Credit dataset [2]. We have studied both these datasets for demonstrating bias with respect to gender as a protected attribute and male as the privileged group.

For the Adult dataset we had taken one group as all female instances having <50k income and other group as all male instances having >=50k income as here privileged group is male and favourable outcome is to have >=50k income. For German credit dataset we had taken one group as all female instances with bad risk value and other group as all male instances with good risk value. In this dataset, male is a privileged group and good credit risk is favourable outcome.

In the experiments performed, we had taken cosine similarity threshold as 0.99 to flag similar samples from both the groups. After that we had ranked all the flagged samples according to count of similar samples from other group and removed top k\% instances. For values of k=1,2 results are illustrated in table 2. We have experimented with various classification algorithm such as XGBoost, LighGBM, Random forest and logistic regression to check the effectiveness of our proposed methodology. For each of the mentioned algorithm we found that the proposed method was able to increase accuracy as well as decrease biasness as shown in table 1. 
\begin{table}
\begin{center}
\begin{tabular}{|c| c| c| c| c|c|c|c|}
    \hline
    \multicolumn{2}{|c|}{} & \multicolumn{3}{|c|}{\textbf{Adult Dataset}} &
    \multicolumn{3}{|c|}{
    \textbf{German Dataset}}\\
    \hline\hline 
    Algorithm & Data & Accuracy & AOD & SPD & Accuracy & AOD & SPD\\ [0.5ex] 
    \hline
      & Raw & 0.84 & 0.19 & 0.18  & 0.73 & 0.12 & -0.03\\
    XGBoost  & \text{1 \% removal} & {\bf0.85} & {\bf-0.01} & {\bf0.09} & 0.74 & -0.05 & 0.04  \\
      & \text{2 \% removal} & 0.85 & -0.03 & 0.10 & {\bf0.76} & {\bf0.01} & {\bf0.04} \\
     \hline
     
       & Raw & 0.84 & 0.19 & 0.18 & 0.72 & 0.07 & 0.04  \\
    LightGBM  & \text{1 \% removal} & 0.84 & -0.05 & 0.09 & {\bf0.74} & -0.06 & 0.01 \\
      & \text{2 \% removal} & {\bf0.85} & {\bf-0.04} & {\bf0.09} & 0.73 & {\bf0.04} & {\bf0.03}  \\
     \hline 
    
       & Raw & 0.82 & 0.19 & 0.18 & 0.70 & 0.03 & 0.05  \\
    Random Forest  & \text{1 \% removal} & 0.83 & {\bf-0.02} & {\bf0.12}  & {\bf0.72} & {\bf0.01} & {\bf0.03}\\
      & \text{2 \% removal} & {\bf0.84} & -0.15 & 0.14 & 0.70 & 0.09 & 0.05  \\
     \hline
     
       & Raw & 0.76 & 0.22 & 0.09 & 0.69 & 0.15 & 0.22  \\
    Logistic Regression  & \text{1 \% removal} & 0.76 & 0.03 & 0.01  & {\bf0.69} & {\bf0.08} & {\bf0.04} \\
      & \text{2 \% removal} & {\bf0.76} & {\bf0.008} & {\bf0.01}   & 0.69 & 0.16 & 0.06 \\
     \hline
     
\end{tabular}
\caption{Fairness and accuracy results on two open source dataset}
\end{center}
\end{table}

\begin{table}
\begin{center}
\begin{tabular}{|c| c| c| c| c|}
    \hline
    \multicolumn{1}{|c|}{} & \multicolumn{2}{|c|}{\textbf{Adult Dataset}} &
    \multicolumn{2}{|c|}{
    \textbf{German Dataset}}\\
    \hline\hline 
     Instance & Male & Female & Male & Female \\
     \hline
    Total & 21790 & 10771 & 690 & 310\\
    \hline
    \text{1 \% removal} & 217 & 107 & 6 & 3\\
    \hline
    \text{2 \% removal} & 435 & 215 & 13 & 6\\
    \hline

\end{tabular}

\caption{Number of samples removed from each dataset}
\end{center}
\end{table}

\section{Conclusion}
From the results mentioned in the previous section, it is evident that after the removal of pseudo label noise i.e., instances with similar features but with different output label, model has become fairer. Pseudo label noise is one of the potential bias imparting instances that might have been generated due to bias at the time of data annotation. As we are removing the biased data from the training set which is responsible for making the model biased, hence a model built on such filtered dataset is unbiased and fair model. We believe that pseudo label noise instances are the ones that are responsible for confusing the model and thereby leading to the distortion and shift of decision boundaries. These instances can be viewed through the lens of label noise as discussed in the survey paper (Frénay et. al. 2014)[16] where authors mentioned that label noise is responsible for the decrease in performance of classifiers and removal of such instances is one of the methods to improve performance of such models.

In this paper, we present an advanced but simplistic data pre-processing and filtering based method to remove bias from the data accompanied by contrasting upswing in the machine learning model performance. It overcomes the limitation of existing methods for bias mitigation at the cost of model accuracy. Promising experimental results on two publicly available open-source datasets makes our research well grounded. 

\section{Future Scope}

As part of the future scope of research, we intend to develop similar in-processing algorithm to remove psuedo label noise. In our current set-up, we have used the cosine similarity metric to find out the close instances. A survey of comprehensive set of similarity metrics and comparison of their impact on bias mitigation can be carried out further. We intend to apply the same methodology to image and other data modalities. Another future direction of exploration would be to try out parallel strategies other than flagging and removing similar instances from training data, such as changing labels of such instances and explore which strategy is suitable for a given context.

\end{document}